\documentclass[conference]{IEEEtran}

\usepackage{cite}
\usepackage{amsmath,amssymb,amsfonts}
\usepackage{bm}
\usepackage{float}
\usepackage{graphicx}
\usepackage{subcaption}
\usepackage{textcomp}
\usepackage{xcolor}
\usepackage{algorithm}
\usepackage{algorithmicx}
\usepackage{algpseudocode}
\usepackage{multicol}
\floatname{algorithm}{Algorithm}

\newcommand{\eq}[1]{Eq.~(#1)}

\newcommand{\ve}[1]{\boldsymbol{\mathbf{#1}}}
\newcommand{\R}{\mathbb{R}}

\def\BibTeX{{\rm B\kern-.05em{\sc i\kern-.025em b}\kern-.08em
    T\kern-.1667em\lower.7ex\hbox{E}\kern-.125emX}}

\begin{document}

\title{Multi-Hypothesis Scan Matching through Clustering}

\author{
\IEEEauthorblockN{Giorgio Iavicoli}
\IEEEauthorblockA{\textit{School of Computer Science}\\
University of Birmingham, B15 2TT, United Kingdom\\
giorgio.iavicoli@gmail.com}
\and
\IEEEauthorblockN{Claudio Zito}
\IEEEauthorblockA{\textit{Technology Innovation Institute} \\
Abu Dhabi, UAE \\
Claudio.Zito@tii.ae}
}

\maketitle

\begin{abstract}
Graph-SLAM is a well-established algorithm for constructing a topological map of the environment while simultaneously attempting the localisation of the robot. It relies on scan matching algorithms to align noisy observations along robot's movements to compute an estimate of the current robot's location. We propose a fundamentally different approach to scan matching tasks to improve the estimation of roto-translation displacements and therefore the performance of the full SLAM algorithm. A Monte-Carlo approach is used to generate weighted hypotheses of the geometrical displacement between two scans, and then we cluster these hypotheses to compute the displacement that results in the best alignment. To cope with clusterization on roto-translations, we propose a novel clustering approach that robustly extends Gaussian Mean-Shift to orientations by factorizing the kernel density over the roto-translation components. We demonstrate the effectiveness of our method in an extensive set of experiments using both synthetic data and the Intel Research Lab's benchmarking datasets. The results confirms that our approach has superior performance in terms of matching accuracy and runtime computation than the state-of-the-art iterative point-based scan matching algorithms.        
\end{abstract}


\section{Introduction}

When a robot need to navigates in an unknown environment, localizing itself becomes an \textit{chicken-and-egg} problem. Algorithmically, a solution can be found with iterative procedures in which an initial coarse estimate is refined when new information is available. Graph-SLAM (Simultaneous Location And Mapping) iteratively constructs and optimises a topological graph of robot's poses in different points in time where edges represent constraints between the poses as roto-translation displacements. Scan matching algorithms estimate this geometrical displacement by aligning the sensory state of previous poses with the current one. Hence robustness and efficiency are of paramount importance for scan matching algorithms.

In this paper we proposed a fundamentally different approach to scan matching tasks by leveraging the power of cluster analysis as a statistical technique to identify possible candidates of roto-translation displacements. We achieve this in a two-step process. First, we generate multiple hypotheses on the transformation between two scans, then we coalesce these transformations into a smaller set of distinct weighted hypotheses. 
Our key insight is that most of the randomly generated hypotheses in stage one will be a good local estimate of the scan alignment, hence the most likely transformation will lay in the region with highest local density of hypotheses. To compute this local densities we propose a modified version of Gaussian Mean-Shift which enables us to robustly clustering rotations as well as translations.

We demonstrate the effectiveness of our approach in two set of experiments. First, we verify that our method is robust to increasingly larger sensor errors in a variety of simulated scenarios. We then demonstrate the superiority of our method in terms of matching accuracy and runtime performance against state-of-the-art iterative scan matching algorithms on the benchmarking datasets proposed by the Intel Research Lab in~\cite{datasets}. 
The rest of the paper is structured as follows. The next section presents a brief overview on scan matching and clustering methods. Section~\ref{sec:approach} introduces the details of our approach while Sec.~\ref{sec:results} presents the experimental evaluations and discusses the results. We conclude the paper with final remarks and an overview of future investigations. 




\section{Background}\label{sec:relatedwork}


\subsection{Scan Matching}

Iterative point-based methods for scan matching tasks have gained a lot of interest over the last two decades~\cite{ICP_survey}; its variants have also been applied successfully to robot manipulation tasks, as in our previous work\cite{bib:zito_2016, bib:zito_w2012, bib:zito_w2013, bib:rosales_2018, bib:zito_2019}. 
The key idea is to repeatedly improve the alignment by forming pairs of corresponding points and using these to estimate the transformation between two scans. The main difference among algorithms of this kind is the metric used to determine the point correspondence.

Iterative Closest Point (ICP) algorithm~\cite{Besl1992ICP}, 
in its simplest version, uses the Euclidean distance for linear and angular displacement. 
One of the drawbacks of the ICP 
is that is unable to obtain a good estimate of the rotation, and instead relies on its convergence over a larger number of iterations.

Iterative Dual Correspondece (IDC) algorithm combines the ICP metric for translations with an Iterative Matching-Range Point (IMRP) for rotations. The IMRP creates correspondences by selecting the point with the most similar range measure. This solves the problem of poor rotational estimates when the scan is taken from the same location but with a different orientation.
In contrast to IDC, our approach models translations and rotations as probability density functions to estimate the most likely observed displacement. Experimental results in Sec.~\ref{sec:results} show that our methods outperforms IDC. 


\begin{figure}[t]
\centerline{\includegraphics[width=0.35\textwidth]{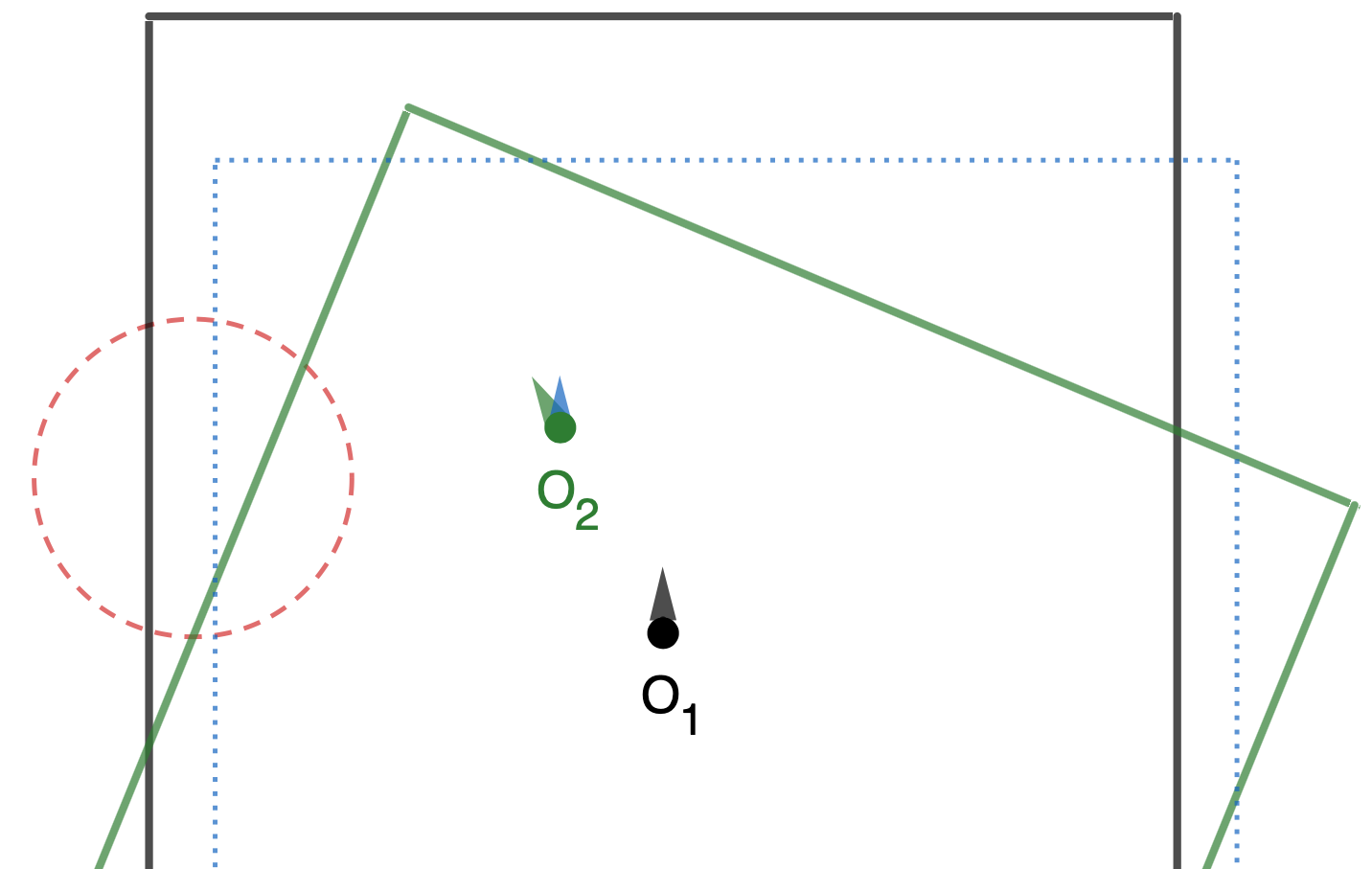}}
\caption{Two range scans taken at two distinct locations: $O_1$ (black) and $O_2$ (green). The blue dotted line represents the latter when corrected for rotation. The red dotted circle represents the selected region of interest in which point $p$ and $q$ are drawn from (See Fig.~\ref{hypgeom}).}
\label{scans}
\end{figure}

\subsection{Clustering}

Clustering techniques are typically based on iterative refinement starting from a set of initial guesses, working in a two step cycle until convergence, where the two steps are point-assignment and parameter-update. 

K-Means (KM)~\cite{KM} is an effective way to partition data into a predetermined number $K$ of clusters. Starting from random guesses for the means, each data point is assigned to the cluster with closest mean and then new means are calculated for each cluster. However the initial random initialization may lead to suboptimal solutions. Convergence speed and accuracy can be improved using K-Means++~\cite{KMpp} where the initial guesses are chosen as far as possible from each other.

In contrast, Mean-Shift (MS)~\cite{MS} uses all the data points as initial guesses, which makes the algorithm more computationally expensive, and the new means are computed by averaging all the points residing withing a predetermined distance or bandwidth. As alternative, weighted Mean-Shift~\cite{WMS} can be considered a hill-climbing algorithm on a density defined by a finite mixture or a kernel density. When the kernel are Gaussians, weighted MS acts as an Expectation-Maximization (EM)~\cite{EM} algorithm which implies that it can converge from almost any starting point~\cite{GMS}. 


Our proposed clustering approach has two main differences. First, part of our initialisation are chosen following the K-Means++ approach, while the rest are randomly selected. Then, the refinement step is similar to a Gaussian MS but with a factorized the kernel function: a two-dimensional Gaussian distribution over the 2D translation displacements, and a von Mises-Fisher distribution which forms a Gaussian-like distribution over rotations~\cite{fisher1953sphere}. Section~\ref{sec:approach} will introduce a more detailed description of our approach.
\section{Proposed Approach}\label{sec:approach}

\begin{figure}[t]
\centerline{\includegraphics[width=0.25\textwidth]{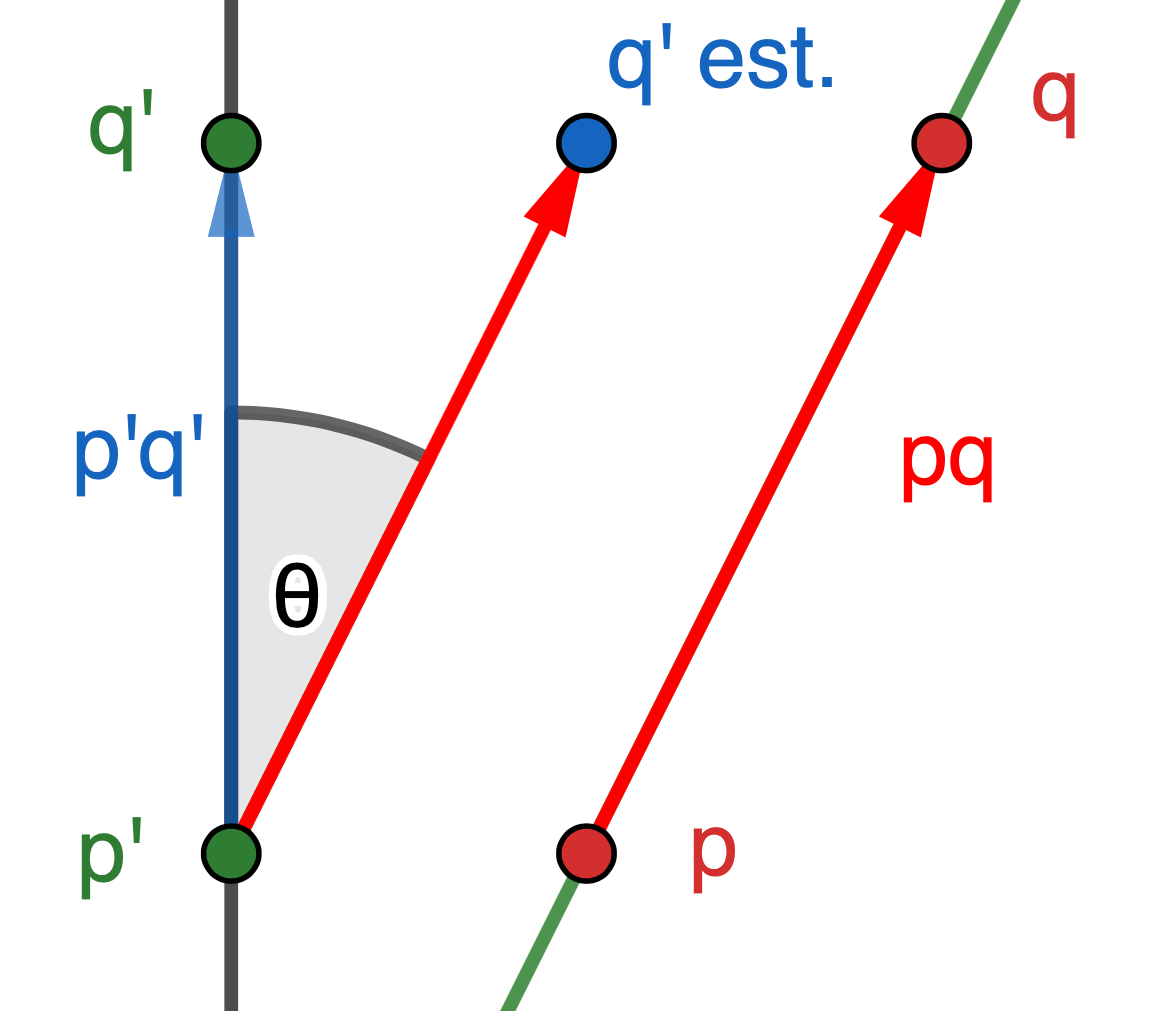}}
\caption{Example of pairwise matching relative to Fig.~\ref{scans}. Point $p'$ is drawn from the KNearestNeighbor of $p$ in $\ve{s}_r$. The vector $\vec{pq}$ is projected in $\ve{s}_r$ and the angle $\theta$ denotes the displacement.}
\label{hypgeom}
\end{figure}

\subsection{Representations}

Without loss of generality, we assume a set of previously estimated poses for the robot as $\hat{X}_t=\{\ve{\hat{x}}_r\mid \ve{\hat{x}}_r=(x_r,y_r,\theta_r)^\top, r\in[0,t-1]\}$. At the beginning of the SLAM algorithm we also assume that the set $\hat{X}$ is composed only of an initial guess $\ve{\hat{x}}_0$. Each pose is represented as a vector $\ve{x}_t=(x_t,y_t,\theta_t)^\top$ where the pair $(x,y)^\top$ represents its position in a two-dimensional space, and $\theta$ its orientation or bearing. For each of these poses we also assume to have a sequence of $N_l$ observations from a laser range sensor as $\ve{z}_t^{(l)}=(d^{(l)},\alpha^{(l)})$ for each ${l\in[0,N_l}]$, where $d^{(l)}\in[0,r_{max}]$ is the distance to an obstacles within its maximum range $r_{max}$ and $\alpha^{(l)}$ is the angle associated to that measurement.
A correspondent position in the Cartesian plane $\ve{p}^{(l)}=(x^{(l)},y^{(l)})^\top \in \mathbb{R}^2$ of a scan point $(d^{(l)},\alpha^{(l)})$  can be computed as follows:
\begin{equation}
    \ve{p}^{(l)} = (x^{(l)},y^{(l)}) = d^{(l)} (\cos{\alpha^{(l)}}, \sin{\alpha^{(l)}})^\top    
\end{equation}


The goal of a scan matching algorithm is to estimate the roto-translational displacement of the robot after its next movement. In other words, at time $t$ we want to estimate the difference between the current (unknown) pose $\ve{x}_t$ and a pose of reference $\ve{\hat{x}}_r$ from our previous estimates, by matching the associated sensory states $\ve{s}_t$ and $\ve{s}_r$. Figure~\ref{scans} and~\ref{hypgeom} show our pairwise matching approach. 


\subsection{Hypotheses Generation}

The goal of the procedure defined in the listing Algorithm 1 is to generate at least a predefined number of hypotheses, $N$, to match the current scan $\ve{s}_t$ against the reference scan $\ve{s}_r$.

\paragraph{Point Pair Matching}

Given a pair of points $(\ve{p},\ve{q})$ both belonging to a scan $\ve{s}$, we denote them as neighbors iff $d_{min}<f_N(\ve{p},\ve{q})<d_{max}$ where
\begin{equation}\label{eq:neighbors}
    f_N(\ve{p},\ve{q})= \sqrt{||\vec{pq}||_2}
\end{equation}
is the magnitude of the vector $\vec{pq}=\ve{q}-\ve{p}$, from point $\ve{p}$ to point $\ve{q}$, and $d_{min}$,$d_{max}\in\R^+$ are user defined parameters. For each point $\ve{p}\in\ve{s}$ we can construct a set of neighbors as 
\begin{equation}\label{eq:neighbor_set}
    Q(\ve{p})=\{\ve{q}\mid \ve{q}\in\ve{s}, d_{min}<f_N(\ve{q},\ve{p})<d_{max}\}.
\end{equation}

Suppose we have a reference range scan $\ve{s}_r$, and one taken at a new pose, $\ve{s}_t$. Assuming a static environment, these only differ by a translation $\Delta T$ and a rotation $\Delta\theta$, and each measurement $\ve{p}\in\ve{s}_r$ is related to the physically corresponding measurement $\ve{p}'\in\ve{s}_t$ by
\begin{equation}
\ve{p}' = R_{\Delta\theta} \times \ve{p} + \Delta T
\label{pointcorr}
\end{equation}
where $R_{\Delta\theta}$ is the rotation matrix 
$\begin{bmatrix} \cos(\Delta \theta) & -\sin(\Delta \theta) \\ \sin(\Delta \theta) & \cos(\Delta \theta) \\ \end{bmatrix}$.

For a random pair of neighboring points in the current scan $\ve{s}_t$, $\ve{p}$ and $\ve{q}$, we can define hypotheses of correspondence to a second pair in the reference scan, $\ve{p}'$ and $\ve{q}'$ in $\ve{s}_r$, by the following characteristics: let $\ve{p}'$ be some K-nearest neighbor of $\ve{p}$ when projected onto the other scan, $\ve{s}_r$, then $\ve{q}'$ will be the closest point to $\hat{\ve{q}'}$, the projection of $\ve{q}$ where point $\ve{p}$ is congruent to point $\ve{p}'$, and we write it as $\ve{p} \cong \ve{p}'$. 
The angle between the vectors $\vec{pq}$ and $\vec{p'q'}$ is an indicator of the rotational displacement between the scans. This is calculated in the algorithm as the difference $\Delta \theta$ between the angle to the $x$-axis of the two vectors.
We can rearrange equation \ref{pointcorr} and using this rotation we can obtain the displacement between $\ve{q}$ and $\ve{q}'$, an estimate of the translation $\Delta T$ between the scans:
\begin{align}\label{eq:finalest}
    \Delta \theta & = \theta' - \theta    \\
    \Delta T & = \ve{q}' - \ve{q} \times R_{\Delta\theta}
\end{align}

\paragraph{Algorithm description}

The algorithm iterates the following until the requested number of hypotheses is generated:
a random point $\ve{p}$ is picked from the current scan, along with the set $Q$ of all points whose distance to $\ve{p}$ is between $d_{min}$ and $d_{max}$. Unless this set is empty, we can proceed by picking a random point $\ve{q}$ from it, and the vector $\vec{pq}$ and its angle to the $x$-axis are calculated.
The K-nearest neighbors of $\ve{p}$ in the other scan $\ve{s}_r$ are computed, and each of these points $\ve{p}'$ in $P'$ makes up the basis for a hypothesis, which is formulated as explained above.
The hypothesis generated is then comprised of three components: the translation $\Delta T$, the rotation $\Delta \theta$, and the contribution factor $\ve{\psi}$. The iteration is then concluded by adding this to the current set of hypotheses.

\paragraph{Contribution factor}

The contribution factor $\ve{\psi}$ is the unit vector representing the direction along which the hypothesis is most reliable.
This is derived from the assumption that data points lie on a continuous surface, and especially when linear such as a wall, the match can give little indication along the vector $\vec{p'q'}$. The direction along which the hypothesis is most likely to be accurate is perpendicular to $\vec{p'q'}$, as the translation is estimated from the distance of $\ve{q'}$ and $\ve{q}$ after accounting for the estimated rotation, $\ve{q}^R$.

\begin{algorithm} [t]
\label{hypgen}
\caption{Hypotheses generation}
\begin{algorithmic}[1]

    \Procedure{GenHyp}{$\ve{s}_t,\ \ve{s}_r,\ N,\ d_{min},\ d_{max},\ K$}
   	\State $H = $ \{\}

    \While{$|H| < N$} 
    	\State $\ve{p} \gets$ random point in $\ve{s}_t$
    	
    	\State Compute $Q(\ve{p})$ using \eq{\ref{eq:neighbor_set}}
    	
    	\If{$Q(\ve{p}) \ne \emptyset$}
    		\State $\ve{q} \sim Q$ ??
    		\State $\vec{pq} = \ve{q} - \ve{p}$
    		\State $\theta \gets$ angle to $x$-axis of $\vec{pq}$
    		
    		\State $P' \gets$ K-NearestNeighbors of $\ve{p}$ in $\ve{s}_r$
    		\ForAll{$\ve{p'}$ in $P'$}

    			\State $\ve{\hat{q}'} \gets$ projection of $\ve{q}$ from  $\ve{p'}$
    			\State $\ve{q'} \gets$ closest point to $\ve{\hat{q}'}$ in $\ve{s}_r$
    			\State $\vec{p'q'} = \ve{q'} - \ve{p'}$
    			\State $\theta \gets$ angle to $x$-axis of $\vec{p'q'}$

    			\State $\Delta \theta = \theta' - \theta$
    			\State $\ve{q}^R =$ Rotate $\ve{q}$ by $\Delta\theta$
    			
    			\State $\Delta T = \ve{q'} - \ve{q}^R $
    			\State $\ve{\psi} = (\cos{(\theta')}, \sin{(\theta')})^\top$
    			\State Append $\langle \Delta T, \Delta \theta, \ve{\psi} \rangle$ to $H$

   			\EndFor
   		\EndIf
    \EndWhile
    
\EndProcedure

\end{algorithmic}
\end{algorithm}

\begin{algorithm} [t]
\label{hypgen}
\caption{Cluster refinement}
\begin{algorithmic}[1]

\Procedure{ClusterRefinement}{$Seeds, Hypotheses$}
    \ForAll{$s$ in $Seeds$}
    	\For{1..$max\ iterations$}
    		\For{$i$ in $1..n$}
			    \State Compute membership weight 
			    \State $w_i = f(\ve{x}_i, \theta_i, \ve{\psi}_i \ |\ \ve{\mu}_s, \Sigma_s, \theta_s, \kappa_s)$
    		\EndFor
    		
    		\State Compute linear mean $\ve{\mu}_s'$ and covariance $\Sigma_s'$
    		\State Compute angular mean $\theta_s'$ and covariance $\kappa_s'$
    	    \State $s = \langle \ve{\mu}_s', \Sigma_s', \theta_s', \kappa_s' \rangle$
    	\EndFor
    	\State Check for convergence
    \EndFor

\EndProcedure

\end{algorithmic}
\end{algorithm}

\subsection{Clustering}

In the second phase the algorithm employs a hybrid clustering technique to combine previously generated hypothesis into few distinct ones. 
The clustering algorithm consists of an initial stage where the initial points for the clusters are chosen, or seeds, and a refinement stage where these are iteratively corrected until convergence. The final set of hypotheses is then obtained by computing the sets of connected components.

\paragraph{Seeding}

Density-based clustering algorithms typically require a set of initial points, also known as seeds.
Careful seeding is very relevant to the proposed clustering method, as this is also density based like K-Means, and therefore requires at least one starting point in a region near the final desired value.
The initial stage uses the same basic principle proposed in K-means++ \cite{carefulseeding}, where a subset of points is generated by iteratively choosing a point at random, with probability proportional to the squared distance to the closest among the previously chosen points. 
It was empirically found that having a fixed proportion of initial points should drawn uniformly at random increased diversity in denser region which would otherwise be underrepresented, and those are also the regions where the best transformation estimate is found.
The ratio in the random selection of $\frac 1 4$ uniform and $\frac 3 4$ distance-weighted seeds was found to provide a good balance.

\paragraph{Iterative refinement}

The iterative refinement shown in Algorithm 2 is very similar to that of the weighted MS algorithm, however it employs a hybrid kernel, which is the product between a two-dimensional Gaussian distribution and a von Mises distribution.

For each cluster in $Seeds$, the algorithm iteratively computes the membership weight of each point by taking into account the linear and angular distances, $m$ and $v$ respectively.

The normal probability density function at a point $\ve{x}$ with mean vector $\ve{\mu}$ and covariance matrix $\Sigma$ in two dimensions is defined as 
\begin{equation}
\mathcal{N}(\ve{x},\ve{\mu},\Sigma) = \frac{\exp{[-\frac{1}{2} (\ve{x} - \ve{\mu})^\top\ \Sigma^{-1}\ (\ve{x} - \ve{\mu})]}} {2\pi \sqrt{det(\Sigma)}}
\label{normdist}
\end{equation}

The descriptive statistics for this distribution is the Mahalanobis distance,
$MD(\ve{x},\ve{\mu}, \Sigma) = \sqrt{(\ve{x} - \ve{\mu})^\top\ \Sigma^{-1}\ (\ve{x} - \ve{\mu})}$.
By ignoring the constant terms that will be accounted for through normalization, we obtain the following:
\begin{equation}
\mathcal{N}(\ve{x},\ve{\mu},\Sigma) \propto \exp{[-\frac{1}{2} MD(\ve{x},\ve{\mu}, \Sigma)^2]}
\label{normprop}
\end{equation}

However, since our hypotheses carry a contribution weight $\psi_i$, this distance should be calculated along that axis. As the contribution vector is a unit vector by construction, we can project the difference $\ve{\mu}_i - \ve{\mu}_s$ onto this factor by multiplication:
\begin{equation}
M(\ve{x}_i, \ve{\psi}_i, \ve{\mu}_s, \Sigma_s) = \sqrt{
    			    (\ve{\psi}_i (\ve{x}_i - \ve{\mu}_s))^\top\ 
    			    \Sigma_s^{-1}\ 
    			    (\ve{\psi}_i (\ve{x}_i - \ve{\mu}_s))
    			  }
\label{mdpsi}
\end{equation}

Similarly, the likelihood of the angular component given the cluster mean is represented as follow
\begin{equation}
\Theta(x\mid\mu,\kappa) = \frac{\exp{[\kappa\ cos(x - \mu)]}} {2\pi I_0(\kappa)}
\label{vonmdist}
\end{equation}

where $\Theta$ is the antipodal von Mises-Fisher distributions which is a very close approximation to the wrapped distribution for circular quantities~\cite{fisher1953sphere}, $I_0(\kappa)$ is the modified Bessel function of order 0\cite{bessel}.
The exponential term of these distributions are
\begin{align}\label{eq:finalest}
    m_i & = M(\ve{x}_i, \ve{\psi}_i, \ve{\mu}_s, \Sigma_s)    \\
    v_i & = \Theta(x\mid\mu,\kappa).
\end{align}

The membership weights for each hypothesis is then the product of the exponential terms found in the probability density functions of the Gaussian and von Mises distributions, $e^{-\frac 1 2 m^2}$ and $e^v$ respectively, which can be simplified as 
\begin{equation}
w_i = e^{-\frac 1 2 m^2_i + v_i}.
\label{membweight}
\end{equation}

\begin{figure*}[ht]
    \centering
    
    \begin{subfigure}[t]{0.3\textwidth}
        \centering
 	 \includegraphics[width=\linewidth]{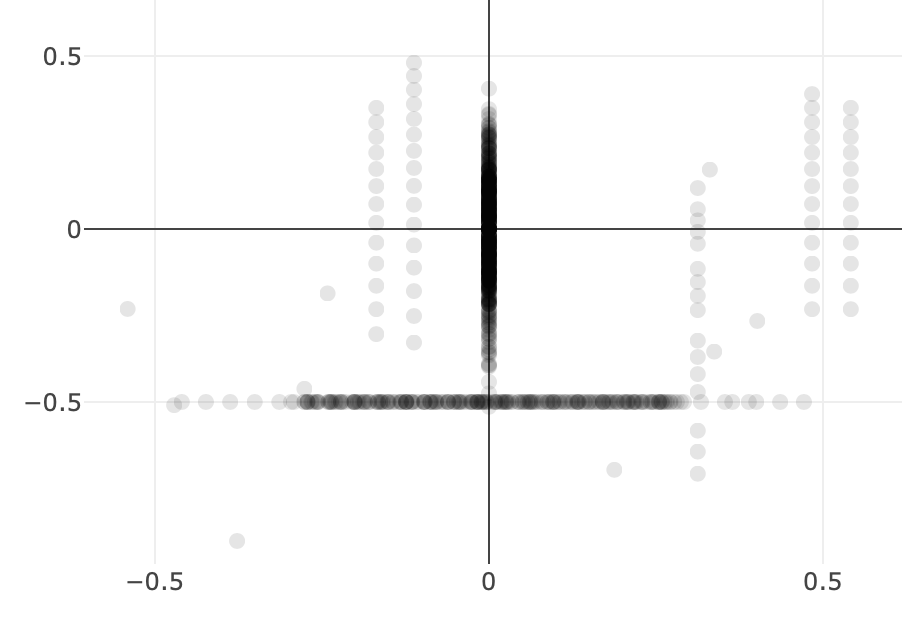}
        \caption{Without noise}
    \end{subfigure}
    \begin{subfigure}[t]{0.36\textwidth}
        \centering
 	 \includegraphics[width=\linewidth]{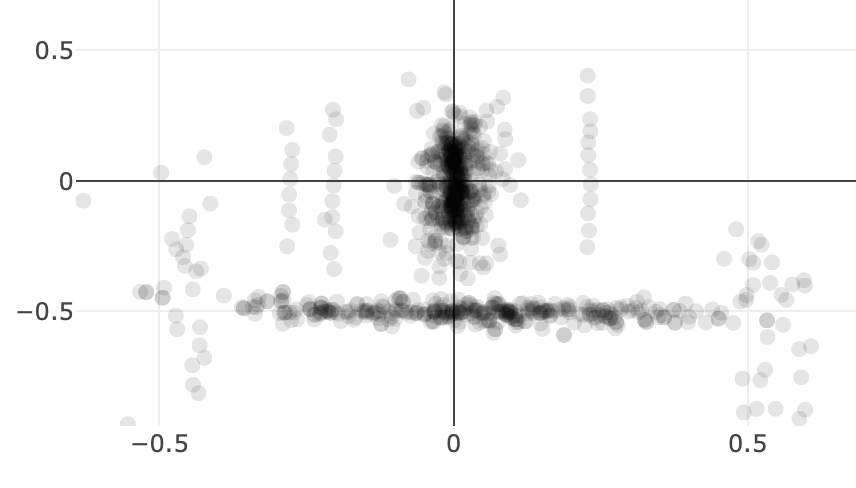}
        \caption{With noise}
    \end{subfigure}
    \begin{subfigure}[t]{0.32\textwidth}
        \centering
 	 \includegraphics[width=\linewidth]{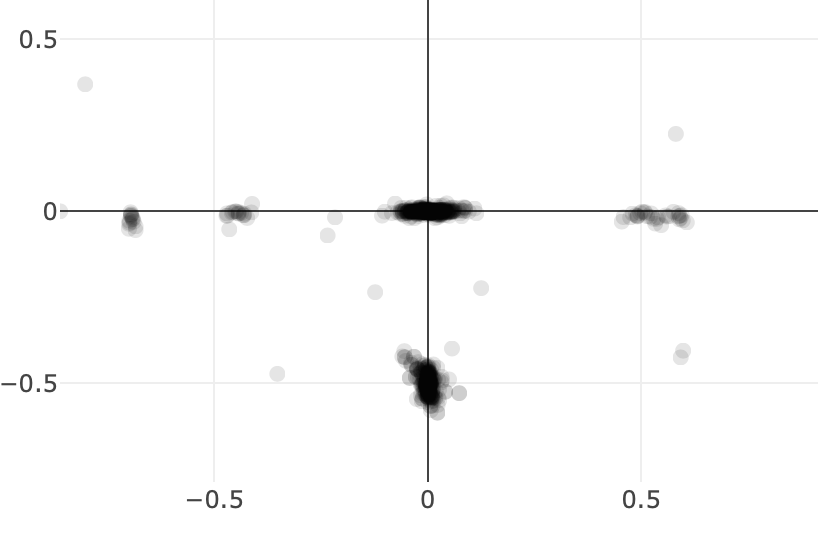}
        \caption{With noise, after weighting by contribution factor}
    \end{subfigure}
    
    \caption{Visualization of translation in hypotheses generated for a transformation ($x=0,\ y=-0.5,\ \theta = 0$)}
    \label{hypviz}
\end{figure*}

\paragraph{Cluster parameters update}
After calculating the membership weight for each hypothesis, the normalizer terms $\eta$ and $\Psi$ are used to scale all successive calculation as if the weights and contribution factors summed to 1:
\noindent\begin{minipage}{.5\linewidth}
\begin{equation}
\eta = \left[\sum_{i=1}^n w_i\right]^{-1}
\label{eta}
\end{equation}
\end{minipage}%
\begin{minipage}{.5\linewidth}
\begin{equation}
\ve{\Psi} = \left[\eta \sum_{i=1}^n w_i \ve{\psi}^{(x)}_i\right]^{-1}
\label{psi}
\end{equation}
\end{minipage}
\vspace{2pt}

The new parameters for the linear components of the cluster, the mean and the covariance, are calculated as follows. 
The new mean $\ve{\mu}_s'$ is the normalized sum of each hypothesis' mean multiplied by its membership and contribution weights, normalized through $\Psi$:
\vspace{-0.2cm}
\begin{equation}
\ve{\mu}_s' = \ve{\Psi} \sum_{i=1}^n w_i\ \ve{\psi}_i\ \ve{\mu}_i 
\label{mean}
\end{equation}
Typically, the covariance is defined as the expected product of the deviations of a set of data from the mean. In this special case, the deviation needs to account for the contribution factor when computing the difference, and the result is normalized with respect to the membership weights through $\eta$.
Therefore, the new covariance $\Sigma_s'$ is calculated as sum of the product of each of these deviations from the new mean $\mu'_s$ multiplied by its membership weight, and then normalized through $\eta$.
\begin{equation}
\Sigma_s' = \eta \sum_{i=1}^n w_i ((\ve{x}_i - \ve{\mu'}_s) \times \ve{\psi}_i)((\ve{x}_i - \ve{\mu'}_s) \times \ve{\psi}_i)^\top 
\label{covariance}
\end{equation}

The new parameters for the angular component of the cluster, $\theta_s'$ and $\kappa_s'$, follow a simpler derivation. The new angle $\theta_s'$ is the arc tangent of the ratio of the weighted sum of the sines and that of the cosines of the angle of each hypothesis. In this case, the normalization factor would appear in both the numerator and the denominator, and can be simplified away.
\begin{equation}
\theta_s' = \arctan{\left(\frac{\sum_{i=1}^n sin(\theta_i)}{\sum_{i=1}^n cos(\theta_i)}\right)}
\label{angmean}
\end{equation}

On the other hand, the new angular concentration factor $\kappa_s'$ is the normalized sum of the cosines of the deviation of each angle form the new mean $\theta_s'$.
\begin{equation}
\kappa_s' = \eta \sum_{i=1}^n w_i \ cos(\theta_i - \theta'_s) 
\label{angmean}
\end{equation}

\paragraph{Convergence detection}
This procedure of updating the weights and recalculating the new parameters can be repeated up to a number of iteration equal to $max\ iterations$; however, the algorithm may converge much earlier.
We can stop it after, for a number of consecutive iterations, the difference (or shift) between the previous and the updated means, $(\ve{\mu}_s, \theta_s)$ and $(\ve{\mu}'_s, \theta'_s)$, is lower than a set threshold, for both translational $D_{thr}$ and rotational $R_{thr}$ components.
\begin{align}\label{eq:convergence}
    |\vec{\ve{\mu}_s\ve{\mu'}_s}| & < D_{thr}   \\
    |\theta_s - \theta'_s| & < R_{thr}
\end{align}

\paragraph{Connected components}
The final stage of the clustering algorithm fuses together hypotheses that lie close to each other.
The algorithm works by checking all pairs of clusters (components) and adding a link (connection) between them if their distance is lower than a set threshold.
This forms a graph between hypotheses where only the ones close enough are connected between them: each of these groups is then merged to form a single hypothesis, which is the mean of the components in the group.
\section{Results}\label{sec:results}

\subsection{Synthetic Environments}
Using data from a synthetic environment allows to evaluate the performance in the absence of unwanted features such as noise in sensor data as well as ensuring the environment is perfectly static. Using a raytrace simulation of a rectangular room, we can see the geometrical and statistical properties of the proposed algorithm.

\paragraph{Hypotheses Generation}

\begin{table}[t]
\centering
\caption{Clustered hypotheses}
 \begin{tabular}{|c|c|c|c|}
    \hline
    X(meters) & Y (meters) & Theta (degrees) & Weight\\
    \hline
0.001 & -0.473 & 1.042 & 0.78\\
\hline
0.004 & -0.003 & 0.669 & 0.22\\
    \hline
  \end{tabular}
  \label{clusres}
\end{table}

Figure \ref{hypviz} shows three different plots of the hypotheses generated by matching two scans of such rectangular environment after the same transformation, in which the robot moves 0.5 meters along the $y$-axis, which is represented vertically in the graphs.
All of the data points are very transparent, so that less-dense areas appear light-gray, whereas darker regions imply a higher concentration of hypotheses, which in some areas is enough that overlapping points make some parts appear as black.

Figure \ref{hypviz} (a) shows the result in absence of sensor noise, where the hypotheses generated by matching walls on the sides (parallel to the motion) generate the vertical group of points that lies at $x=0$ and along the $y$-axis.
Conversely, the horizontal line at $y=-0.5$ and that spans the $x$-axis is generated by the matches of the walls in front and behind (perpendicular to the direction of motion) the sensor, and therefore are the ones informative of the translation undertaken.

Figure \ref{hypviz} (b) shows hypotheses generated in presence of sensor with noise comparable to that of sensor mounted on small robots \cite{zeromean}. The two lines described above are still discernible thanks to the linearity of the simple environment.

Figure \ref{hypviz} (c) on the other hand, shows the same data plotted in (b), with the difference that these are now weighted by the contribution factors, which indicate the quantity of information given by measurements in each direction.
The hypotheses that originally generated by walls on the sides, which are those that were $x=0$ and varying in $y$-coordinates by approximately ±0.2m, now reside much closer to (0,0) as these contained no information in the $y$-direction.
Similarly, the data points generated by walls in the front and back, which are those that were at $y=-0.5$ and varying in $x$-coordinates by approximately ±0.2m, now reside very tightly around (0,-0.5) as these contained no information in the $x$-direction.
The other data points that were previously of a very light-grey sequence of points are now also condensed tightly around a single point, however a closer look at the plot shows that these regions are nowhere near as dense as the ones containing the true solution.

\paragraph{Cluster Analysis}

\begin{table}[t]
\centering
\caption{Relative transformations at each time step}
  \begin{tabular}{|c|c|c|c|}
    \hline
    Time step & X(meters) & Y (meters) & Theta (degrees)\\
    \hline
1 & 0 & 0 & 0 \\
    \hline
2 & 0 & -0.5 & 0 \\
    \hline
3 & 0 & 0 & 45 \\
    \hline
4 & -0.35 & 0 & 0 \\
    \hline
5 & 0 & 0 & 45 \\
    \hline
6 & -0.5 & -0.25 & 0 \\
    \hline
  \end{tabular}
  \label{traj}
\end{table}

\begin{figure}[t]
    \centering
    \includegraphics[width=0.7\linewidth]{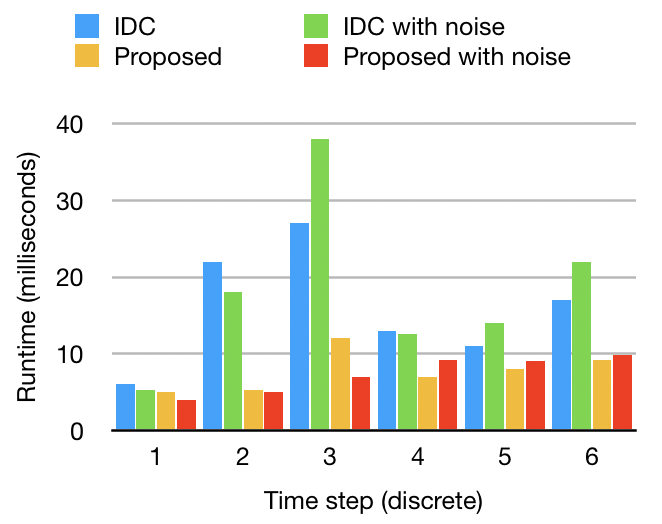}
    \caption{Average time per match on the simulated environment}
    \label{runtime}
\end{figure}

It is now care of the clustering component of the proposed solution to identify the different groups of hypotheses and deduce the importance of each group: the two clusters that are found in the case of the transformation with noise in the sensor data are reported in Table \ref{clusres} along with their weight. 
The first cluster reported is very close to the actual transformation undertaken, which is (0 meters, -0.5 meters, and 0 degrees). The second cluster represents the case where the hypothesis is derived only by the matches along the side walls, ignoring that of the front and back: this would be the case if the measurements that are used to obtain the first cluster were to be outliers, and therefore represents a plausible second hypothesis.
However, when these are weighted based on Average Squared Residual, the former is found to carry a weight of 0.78, or 78\%, while the second has only 0.22, or 22\%, which imply that the first hypothesis is the most likely one.

\paragraph{Runtime on simulated environment}
The graphs in figure \ref{runtime} show the runtime performance in different scenarios, whose sequence of transformations are shown in Table \ref{traj}. It can be seen that the proposed algorithm has better performance, and is generally independent of the underlying transformation. IDC on the other hand is very sensible to these changes, as can be noted in the spike at time step 2 and 3.

\subsection{Public Datasets}

In order to verify the behavior of the algorithm in a real-world scenario, it will be tested on a widely used dataset, which is that of the Intel Research Lab in Seattle, available at \cite{datasets}. This has been used in a variety of previous publications \cite{Intel1} \cite{Intel5} \cite{Intel3} \cite{Intel4} \cite{Intel2}.

\paragraph{RMS Error}

\begin{figure}[t]
    \centering
    \includegraphics[width=\linewidth]{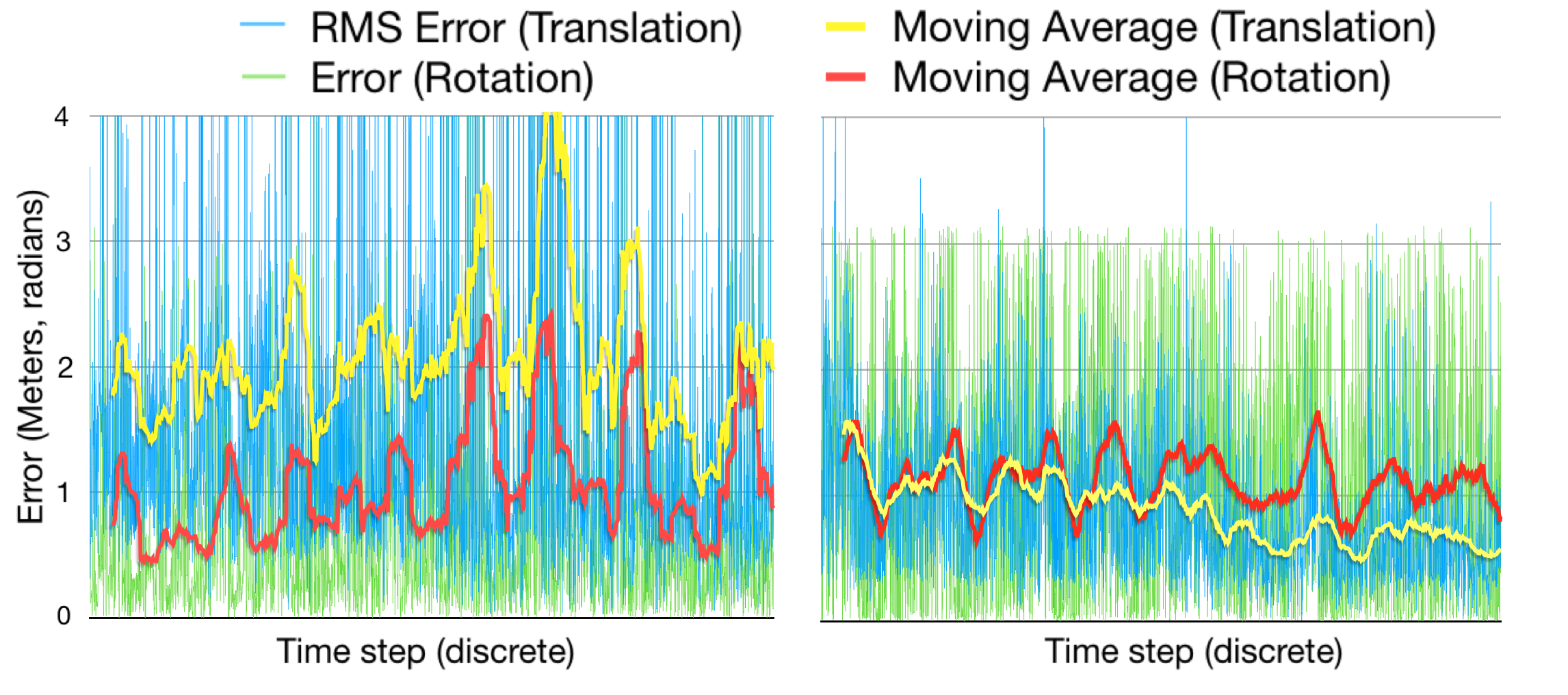}
    \caption{RMS Error over all pairwise matches, IDC (left) and proposed approach (right)}
    \label{RMSE}
\end{figure}

The graphs presented in Fig.~\ref{RMSE} show the RMS Error in translation in blue, and the error in rotation in green, along with their moving averages shown in yellow and red respectively.
The average RMS error in translation over the whole dataset is 2.08 meters for IDC, and 0.894 meters for the proposed approach, which means that IDC is vastly outperformed in this metric.
On the contrary, the average error in rotation is 1.06 radians---or around 61 degrees---for IDC, and 1.12 radians---or around 65 degrees---for the proposed approach, which means that IDC slightly outperforms the proposed approach in this metric.

    

\paragraph{Cumulative Probability of Error}

\begin{figure}[t]
    \centering
    \begin{subfigure}[t]{0.24\textwidth}
        \centering
 	 \includegraphics[width=\linewidth]{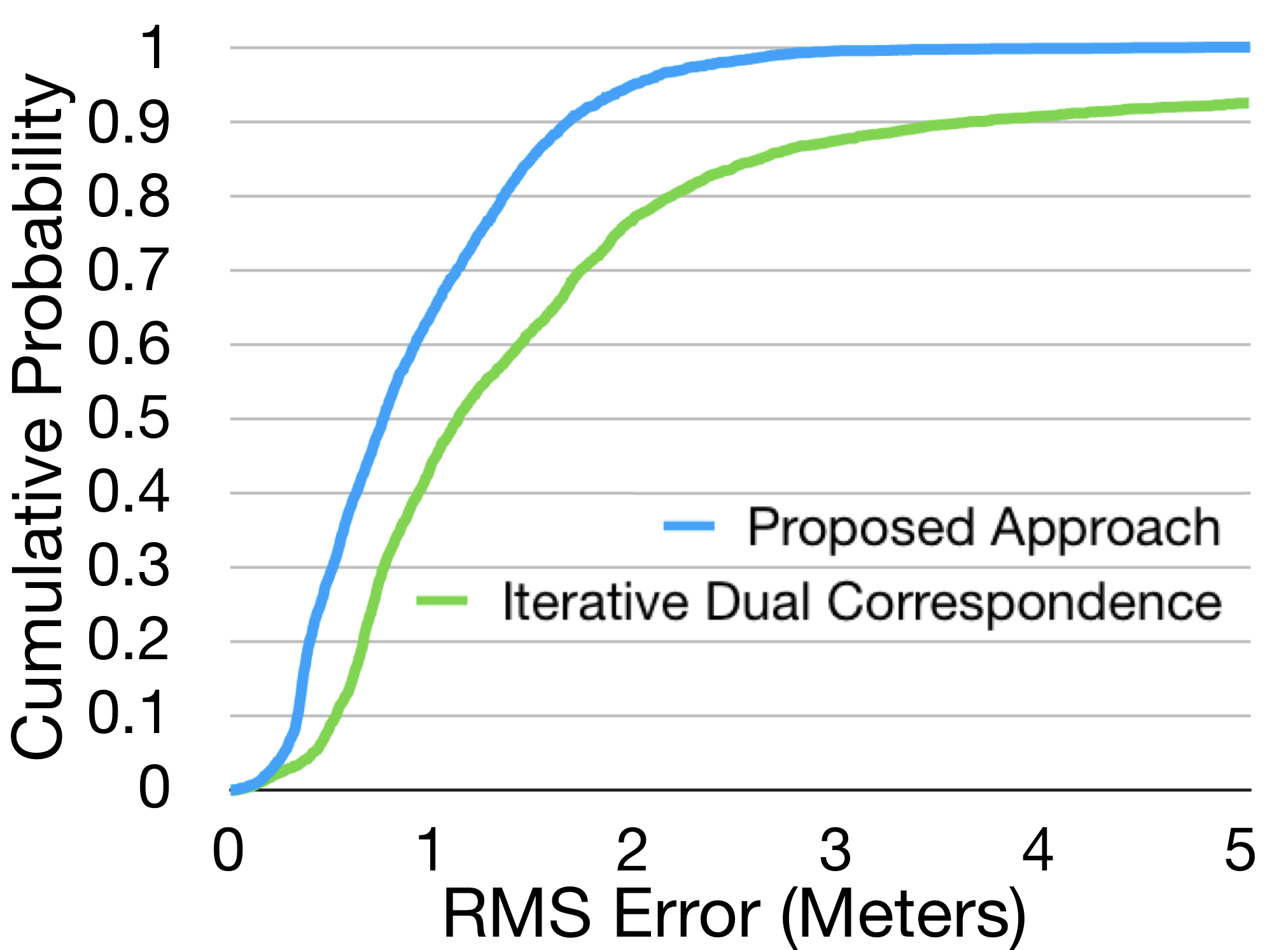}
        \caption{Translation}
    \end{subfigure}
    \begin{subfigure}[t]{0.24\textwidth}
        \centering
 	 \includegraphics[width=\linewidth]{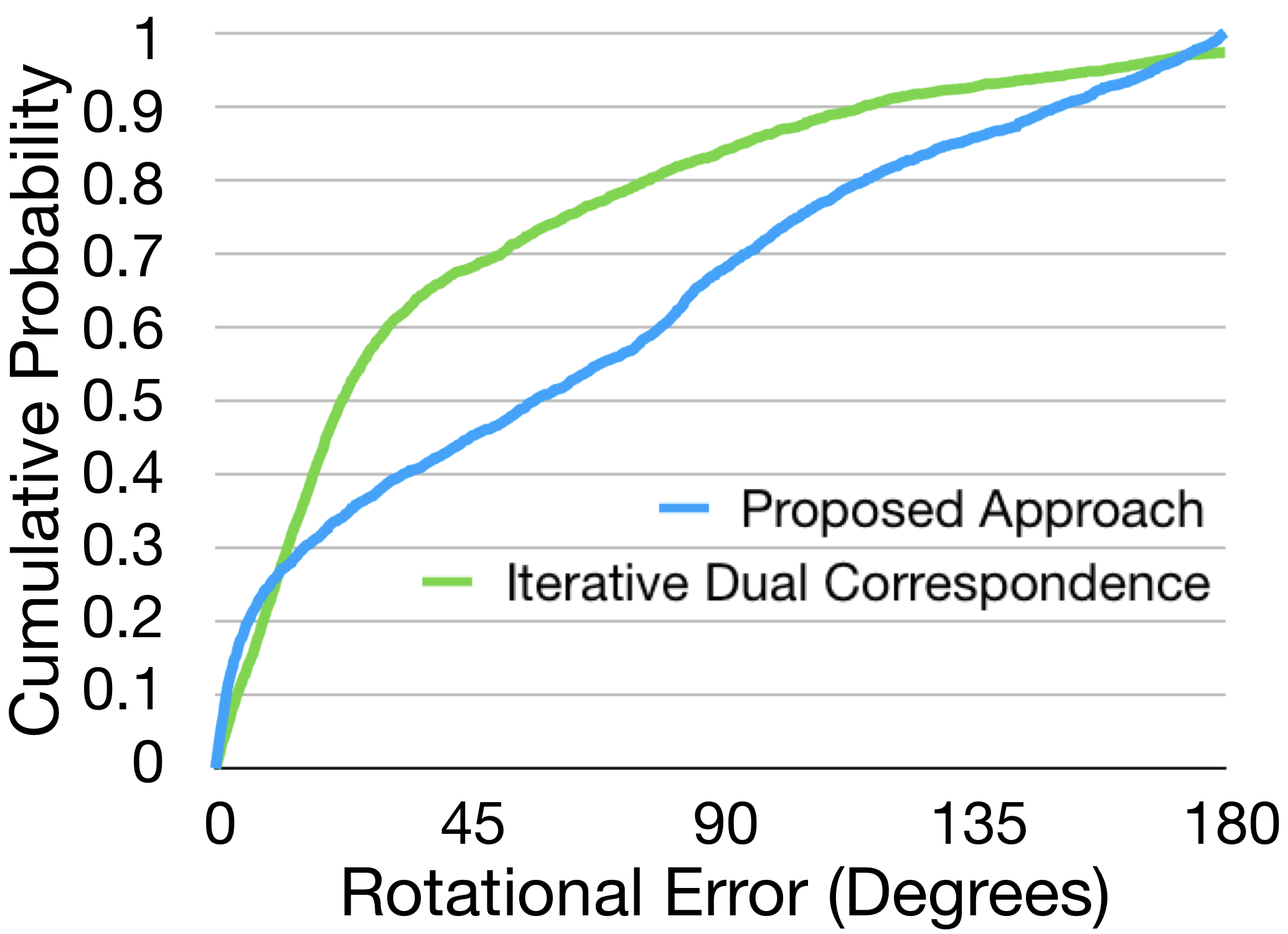}
        \caption{Rotation}
    \end{subfigure}
    
    \caption{Cumulative probability of error distribution}
    \label{CumProb}
\end{figure}

On the other hand, Fig.~\ref{CumProb} shows the cumulative probability of the error being lower.
For example, in \ref{CumProb} (a) we can see that the frequency of the translation error for the proposed approach being lower than 1 meter is approximately 0.65, or for 65\% of the matches. 
Conversely, the frequency for the IDC on the same error level is approximately 0.45, meaning that only 45\% of the matches have an error lower than 1 meter.
It can be then inferred that the proposed approach provides with better estimates of the translation component.
For the rotational component, IDC tends to provide better accuracy for middle range rotational errors, but the runtime performance is much worst. On average, IDC takes 10.97 milliseconds per pair match, whereas the proposed approach averages slightly less than half, at 4.84 milliseconds. 

\paragraph{Runtime on datasets}
Both of these algorithm would be able to process data within the timeframe that separates sensor updates; however, these benchmarks are run on an Intel i7-7700HQ running at 2.8GHz. On smaller mobile platforms employing slower, less power-hungry processors, the performance of these may be very different, and therefore having a lower runtime is still a desirable property for this algorithm.

\section{Conclusion}

In this paper we have presented a new perspective on scan matching for the estimation of roto-translation displacements, which shows the potential use of clustering techniques in SLAM problems. We have shown that is possible to start with a set of randomly generated roto-translation hypotheses and infer the structure of these through cluster analysis by finding the prevalent mode in the data. The proposed clustering algorithm exploits an Expectation-Maximization technique for a Gaussian Mixed Model, in which selected hypotheses becomes our kernel centres. We enable clustering to be robustly applied to both translation and rotational displacements by factorising the kernels.

We have demonstrated the benefits of our approach in an extensive set of experiments. We have shown that our approach is robust to increasingly large sensor errors in a variety of synthetic scenarios, as well as demonstrated improved accuracy and runtime performance against the Iterative Dual Point algorithm on the Intel Research Lab's benchmarking datasets of real robot's scans. 

A drawback of the algorithm is that entirely relies on the alignment of the scans. As future investigation we would like to take advantage of odometry information, if available, to reject hypotheses that lie too far from the true potential region. Additionally, we will investigate a possible extension of this method to 3D scans to enable localisation via depth images. 


\bibliographystyle{plain}
\bibliography{biblio.bib}

\end{document}